**Chapter 6**

# Sequential pattern mining in educational data: the application context, potential, strengths, and limitations

**Yingbin Zhang**[1], **Luc Paquette**[2]

[1]Affiliation: The Department of Curriculum and Instruction, College of Education, University of Illinois at Urbana-Champaign
Address: 1310 S Sixth Street, Champaign, IL 61820, United States
Email: yingbinzhang25@hotmail.com
[2]Affiliation: The Department of Curriculum and Instruction, College of Education, University of Illinois at Urbana-Champaign
Address: 1310 S Sixth Street, Champaign, IL 61820, United States
Email: lpaq@illinois.edu

**Abstract**. Increasingly, researchers have suggested the benefits of temporal analysis to improve our understanding of the learning process. Sequential pattern mining (SPM), as a pattern recognition technique, has the potential to reveal the temporal aspects of learning and can be a valuable tool in educational data science. However, its potential is not well understood and exploited. This chapter addresses this gap by reviewing work that utilizes sequential pattern mining in educational contexts. We identify that SPM is suitable for mining learning behaviors, analyzing and enriching educational theories, evaluating the efficacy of instructional interventions, generating features for prediction models, and building educational recommender systems. SPM can contribute to these purposes by discovering similarities and differences in learners' activities and revealing the temporal change in learning behaviors. As a sequential analysis method, SPM can reveal unique insights about learning processes and be powerful for self-regulated learning research. It is more flexible in capturing the relative arrangement of learning events than the other sequential analysis methods. Future research may improve its utility in educational data science by developing tools for counting pattern occurrences as well as identifying and removing unreliable patterns. Future work needs to establish a systematic guideline for data preprocessing, parameter setting, and interpreting sequential patterns.

**Keywords**: sequential pattern mining, pattern recognition, educational data science, learning process, temporal analysis, educational technology

**Abbreviations**

| | |
|---|---|
| EDS | Educational data science |
| LSA | Las-sequential analysis |
| MOOC | Massive Open Online Course |
| SPM | Sequential pattern mining |
| SRL | Self-regulated learning |

## 6.1 Introduction

Learning is the acquisition process of knowledge and skills, which takes time to manifest in behavioral changes (Soderstrom & Bjork, 2015). Thus, temporality is innate in learning, and an increasing number of researchers have suggested the temporal analysis of learning (Hadwin, 2021; Knight & Wise et al., 2017; Molenaar & Wise, 2022; Reimann, 2009). In particular, Molenaar and Wise (2022) summarized four distinctive values of temporal analysis: detecting transitions between learning events, identifying variation in learning processes, explaining variation in learning outcomes, and boosting the emergence of new questions.

In general, two types of temporal properties of learning have been formed: the passage of time and the order in time (Molenaar & Wise, 2022). The passage of time concerns when, how often, or how long learning events of interest occur. A limitation of this temporal property is that it omits events before and after the events of interest, i.e., the contextual information (Winne, 2010). By contrast, the order in time addresses this limitation by focusing on the relative arrangement of learning events, for example, a sequence of events indicating that a learner reads relevant material after viewing the results of a quiz. The relative arrangement of learning events has attracted increasing interest (Caglar Ozhan et al., 2022; Emara et al., 2018; Liu & Israel, 2022; Mishra et al., 2021; Zheng et al., 2022; Zhu et al., 2019). Sequential pattern mining (SPM), as a pattern recognition technique, is powerful for uncovering this temporal property of learning (Moon & Liu, 2019; Van Laer & Elen, 2018). Particularly, SPM can uncover hidden patterns of ordered events that have interesting properties, e.g., frequent in high-performing learners but infrequent in low-performing learners.

This chapter aims to assist educational researchers in understanding the basics of SPM, its potential in Educational Data Science (EDS), and how to achieve the potential. Section 6.1 introduces SPM and several terminologies. Section 6.2 highlights how SPM can be applied to educational data from various channels for various purposes. Section 6.3 uses example studies conducted on a computer-based learning environment to illustrate general ways in which SPM can contribute to theories of engagement and learning. Section 6.4 discusses the strengths, limitations, and future directions of applying SPM to educational data.

### 6.1.1 Sequential pattern mining (SPM)

SPM was first proposed by Agrawal and Srikant (1995) as the problem of finding interesting subsequences in a sequence dataset. The interestingness of a subsequence can be defined in various ways, such as its frequency and length. Mathematically, a sequence can be denoted by $\{i_1, i_2, ..., i_n\}$, where $i_j$ is an itemset. An itemset is a non-empty set of items, which can be various things. For example, in educational data, an item may be an action that a student executes in a learning management system, such as downloading a lecture note and submitting an assignment solution. Agrawal and Srikant (1995) used SPM in the context of customer transactions, where an item is a product, and an itemset is the set of products purchased in one transaction. All of the customer transactions form a sequence. Table 6.1 provides an example of sequences for hypothetical customers. In the table, customer 1's sequence has two transactions. The first itemset contains two items, a and b, indicating that products a and b were purchased in transaction 1, while transaction 2 indicates that c was purchased. The order of items in an itemset is not meaningful.

**Table 6.1**. Examples of transaction sequences

| Customer ID | Customer sequence (Transaction history) |
|---|---|
| 1 | $< \{a, b\}, \{c\}, \{d\} >$ |
| 2 | $< \{a, c\} >$ |
| 3 | $< \{b\}, \{c, d, e\}, \{f, b\}, \{c\} >$ |
| 4 | $< \{b, c\}, \{d\}, \{g\} >$ |

*Note*. A pair of curly braces represents one transaction. Items in a pair of curly braces are products purchased in the transaction.

A sequence $\{i_1, i_2, ..., i_n\}$ contains a subsequence $\{p_1, p_2, ..., p_m\}$ if there are integers $k_1 < k_2 < \cdots < k_n$ such that $\{p_1 \subseteq i_{k_1}, p_2 \subseteq i_{k_2}, ..., p_m \subseteq i_{k_n}\}$. For example, sequences 1 and 3 in Table 6.1 contain the subsequence $< \{b\}, \{c\} >$, but sequences 2 and 4 do not. Note that these integers do not need to be consecutive, i.e., $k_{j+1} - k_j$ can be larger than 1. $k_{j+1} - k_j$ is named the gap between $i_{k_j}$ and $i_{k_{j+1}}$. Many SPM algorithms allow users to set the maximum gap to reduce noisy patterns (patterns that are identified as frequent due to random error) and limit the number of patterns returned (Srikant & Agrawal, 1996; Zaki, 2000). For example, if the maximum gap is 1, the algorithm will infer that customer 1's sequence does not contain $< \{b\}, \{d\} >$ because the gap between b and d in this sequence is 2. It is especially important to remove noisy patterns because they cannot provide reliable information about the processes of interest (e.g., learning processes) that generated the sequences.

A customer supports a subsequence if the customer's sequence contains the subsequence (Agrawal & Srikant, 1995). The support value of a subsequence is defined as the proportion of sequences that contain this subsequence. If the support value of a subsequence is no less than a pre-specified threshold (typically called the minimum support), this subsequence is a frequent sequential pattern. For example, let us set the support threshold as 0.5. In Table 6.1, the support values of $< \{b\}, \{c\} >$, $< \{b\}, \{d\} >$, $< \{c\}, \{d\} >$ are 0.5, 0.75, and 0.5, respectively, so these subsequences are frequent sequential patterns. In contrast, $< \{a\}, \{c\} >$ is not a frequent sequential pattern because its support is $0.25 < 0.5$. Similarly to the use of a maximum gap, selecting appropriate minimum support helps reduce noisy patterns as the more sequences that contain a specific pattern, the less likely that this pattern occurs due to random error.

Another value frequently used by educational research is the instance value (Lo et al., 2008) which refers to the number of occurrences of a sequential pattern in a sequence. For example, if we do not consider the maximum gap, the instance values of $< \{b\}, \{c\} >$ in sequences 1 to 4 are one, zero, two, and one, respectively. Sequence 3 has two occurrences of $< \{b\}, \{c\} >$ rather than three because occurrences are usually counted according to the non-overlapping rule: an item(set) cannot occur at the same position in two instances of a pattern (Ding et al., 2009; Wu et al., 2020). Some studies have used F-support to refer to the proportion of sequences that contain a pattern and I-support to refer to the occurrences of a pattern (e.g., Liu & Israel, 2022; Mudrick et al., 2019).

### 6.1.1.1 SPM algorithms

Agrawal and Srikant (1995) developed three SPM algorithms: AprioriAll, AprioriSome, and DynamicSome. Subsequent studies have proposed better-performing algorithms that allow users to filter patterns using constraints beyond the support value. For example, the Generalized Sequential Patterns (GSP) algorithm allows users to restrict the maximal and minimum gaps between itemsets (Srikant & Agrawal, 1996). Various algorithms have been applied to educational data, such as GSP, constrained Sequential PAttern Discovery using Equivalence classes (cSPADE; Zaki, 2000), Prefix-projected Sequential pattern mining (PrefixSpan; Jian et al., 2004), Sequential PAttern Mining (SPAM; Ayres et al., 2002), Protein Features EXtractor using SPAM (Pex-SPAM; Ho et al., 2005). For readers interested in the details of SPM algorithms, we recommend the review articles by Mooney and Roddick (2013) and Fournier-Viger et al. (2017). It is noteworthy that all of the different algorithms will produce the same set of sequential patterns, given the same constraints (Fournier-Viger et al. 2017). Thus, researchers can use any SPM algorithms that they are familiar with unless the algorithms do not allow constraints in patterns that may have a critical influence on the results.

For readers new to SPM, we recommend cSPADE (Zaki, 2000). It is a fast algorithm, although not the fastest. It allows various constraints on patterns and has been implemented in Python and R[1], which are tools that EDS researchers may be most familiar with. The next section uses synthetic behavioral sequences to demonstrate the application of cSPADE.

### 6.1.1.2 An example of applying SPM to synthetic behavioral sequences

Table 6.2 displays the four synthetic sequences on which cSPADE was applied. We used a maximum gap of one and minimum support of 0.5 without other constraints. Note that we decided these constraints arbitrarily for the purpose of this example. Readers should set the constraints according to their research context and purposes. The appendix presents the Python and R code that we used in this example. In both cases, the code first creates the sequences and formats the data. Then, it applies the cSPADE algorithm. Finally, it outputs the identified patterns.

While cSPADE regards individual events as patterns, we discarded these single-event patterns because they do not capture the relative arrangement of events. cSPADE identified four frequent patterns with a support value greater or equal to the minimum support (Table 6.3). *Attempt → note*, *note → attempt*, and *attempt → note → attempt* had a support value of 0.5, meaning that two students (S1 and S3 in Table 6.2) contained these patterns. *Read → attempt* had a support value of 0.75, meaning that three students showed this pattern. Student S1 did not show this pattern because their read action (i1) was followed by a *hint* action (i2). However, since S1's third action (i3) is an attempt, if we used a maximum gap of two, cSPADE would have considered that S1 showed the *read → attempt* pattern because the second action after *read* (i1) was *attempt* (i3).

This section showed a simple example of the application of SPM to student behavioral sequences. Subsequent sections will discuss more advanced applications of SPM: how to combine SPM with theoretical considerations and other EDS methods to analyze educational data in various modes for various research purposes.

**Table 6.2**. Four synthetic behavioral sequences

| Event ID Student | i1 | i2 | i3 | i4 | i5 | i6 |
|---|---|---|---|---|---|---|
| S1 | read | hint | attempt | note | attempt | attempt |
| S2 | read | attempt | - | - | - | - |
| S3 | hint | read | attempt | note | attempt | - |
| S4 | hint | note | read | attempt | - | - |

---

[1] The implementation of cSPADE in Python and R: https://pypi.org/project/pycspade/, https://CRAN.R-project.org/package=arulesSequences

**Table 6.3**. Frequent patterns in the synthetic behavioral sequences

| Patterns | Support |
|---|---|
| *attempt → note* | 0.50 |
| *note → attempt* | 0.50 |
| *read → attempt* | 0.75 |
| *attempt → note → attempt* | 0.50 |

## 6.2 What modes of educational data and research purposes is SPM applicable to?

While SPM may be less popular than other EDS methods, such as predictive modeling and clustering, it has been applied for a variety of educational purposes using data from different modes across a wide range of educational settings. In particular, computer-based educational systems, such as Learning Management Systems (LMS), Massive Open Online Courses (MOOC), Intelligent Tutoring Systems (ITS), computer-supported collaborative learning environments, educational games, and course enrollment systems, are the most common educational settings for the application of SPMs.

This section discusses some of the most common data modes and research purposes for the application of SPM. These two factors were selected as they are the main factors that determine the procedures of SPM. We discuss the two factors separately because there is no consistent one-to-one match between data modes and purposes. For instance, researchers may apply SPM to students' event logs from an educational game for feature engineering or understanding how students interact with the environment. Similarly, SPM may be used to analyze students' verbal conversations rather than the event logs to reveal how students collaborate in a computer-supported collaborative environment.

### 6.2.1 Data modes

In general, any sequence dataset where sequences are composed of ordered itemsets can be the input of the SPM analysis. This allows researchers to apply SPM to educational data from various channels. It is important to note that itemsets for SPM must be composed of categorical variables. Thus, SPM cannot be applied to sequences of numerical variables unless researchers code the variables into discrete categories.

### 6.2.1.1 Event logs

Event logs provide detailed traces of students' interaction with digital learning environments (Paquette & Bosch, 2020; Winne, 2020; Zhou et al., 2010). Such logs can be collected unobtrusively and without the need for specialized equipment (e.g., microphones, webcam, motion sensor, etc.) and are easily accessible. For these reasons, event logs are the most common type of educational data to which SPM has been applied. For example, Jiang et al. (2015) applied SPM to students' event logs in Virtual Performance Assessments, a computer-based environment that assessed science inquiry skills. In this context, SPM was used to identify how students without experience with the environment showed more sequential patterns indicative of exploration behaviors than experienced students. Emara et al. (2018) compared students' behaviors between working individually and collaboratively in a computer-based science learning environment. Those working individually showed sequential patterns that mainly consisted of reading behaviors, while those working collaboratively showed that patterns consisted of more diverse activities. Kia et al. (2020) mined students' interactions with a learning dashboard. They found that low-performing students showed more frequent sequential behavioral patterns in the learning dashboard compared to other students.

An itemset in event logs usually has only one event because students rarely initiate two actions simultaneously in a learning environment. Note that raw event logs may contain noises and be at a small grain size (Kinnebrew et al., 2013; Zhou et al., 2010). Researchers need to preprocess the raw event logs before using them as input for SPM algorithms. Otherwise, the algorithm may generate many meaningless or uninterpretable sequential patterns. Five main ways of preprocessing event log data for SPM emerge in existing studies: filtering, collapsing, contextualizing, abstraction, and breaking. The choice of which preprocessing methods to use (and how to apply them) relies on the purpose of the SPM analysis, as preprocessing will significantly impact the nature of the patterns that SPM identifies as interesting.

Filtering refers to removing events that may be necessary for students' interactions with a learning environment but irrelevant or meaningless in the context of the studied educational concept (Kinnebrew et al., 2013). For example, a study may focus on students' behaviors after taking a quiz in a MOOC course. The MOOC platform may automatically display the quiz results and record the action of viewing quiz results. In this case, the action of viewing quiz results may not be meaningful because it is required by the platform, and this action may need to be removed.

Collapsing represents condensing consecutive and qualitatively similar events as one event. Collapsing events reduces redundant patterns. For instance, Kinnebrew et al. (2013) condensed a chain of consecutive *reading page* events into a *reading page-MULT* event. In this way, chains of reading events with different lengths will be regarded as the same event (a *reading page-MULT* event) by SPM algorithms, and patterns only differing in the chain length would be considered as the same pattern. For example, both *view quiz results -> reading page -> reading page* and *view quiz results -> reading page -> reading page -> reading page* would be considered as *view quiz results -> reading page-MULT*. Meanwhile, the suffix *MULT* keeps the information indicating that multiple events occurred and distinguishes a chain of reading events from a single reading event.

Contextualizing refers to labeling events with contextual information, such as the duration of an event. For instance, Emara et al. (2018) added *long* or *short* suffixes to *reading page* events based on whether the read events were longer than three seconds. This distinction was meaningful in that context because short reading might indicate that students were skimming pages, while long reading was more likely to indicate that students were trying to understand the page content.

Abstraction refers to translating raw event subsequences into more abstract behaviors. This can be achieved by matching raw event subsequences with pre-specified patterns representing higher-order behavior. For instance, in Zhou et al.'s (2010) work, a subsequence, *selecting content -> opening a note window -> choosing the option "critique note" -> filling information -> closing the note window*, was coded as the behavior of *making a critique note*. Abstraction can be especially useful in situations where a raw event in the logfile rarely represents a meaningful action.

Breaking refers to breaking raw sequences based on research interests. For instance, a group's event sequence in a collaborative learning environment may be split into student-based sequences where each sequence contains ordered events from a group member (Perera et al., 2009). Sequential patterns in student-based sequences reflect common behaviors within this group. A student-based sequence may be further split into session-based or day-based sequences, with each containing events from a single learning session or day (Kang & Liu, 2020; Zhou et al., 2010). Comparing discovered sequential patterns from different sessions or days may reveal how learning behaviors evolve across sessions and days.

### 6.2.1.2 Discourse data

SPM has been applied to the analysis of different forms of discourse data, including students' verbal conversations (Swiecki et al., 2019), online chat messages (Zheng et al., 2019; Zhu et al., 2019), and forum posts (Chen et al., 2017). Discourse data need to be coded into a set of categories based on coding frameworks, where the codes represent abstract behaviors. In this case, an itemset is a code, and a sequence is composed of codes ordered in time. For example, work by Zheng et al. (2019) coded chat messages into eight categories based on the theory of self-regulated learning (SRL) and socially shared regulated learning (SSRL): SRL- and SSRL-task analysis, SRL- and SSRL-planning, SRL- and SSRL-elaborating, as well as SRL- and SSRL-monitoring. Such categorization allowed them to use SPM to explore patterns of socially shared regulation from chat logs. Similarly, Chen et al. (2017) studied schemes of knowledge-building discourse using SPM to analyze forum posts coded into six categories: questioning, theorizing, obtaining information, working with information, syntheses and analogies, and supporting discussion.

### 6.2.1.3 Resource accessing traces

For data from this channel, a sequence consists of learning resources ordered based on the timestamp at which a student accessed it. A learning resource may be a course (Elbadrawy & Karypis, 2019; Jin et al., 2017), a lecture video (Bhatt et al., 2018; Wang et al., 2019), a book (Anwar et al., 2020; Sitanggang et al., 2010), a web page (Chen & Wang, 2020; Chen et al., 2014), etc. The application of SPM to these data has mainly been in the context of building recommender systems, providing suggestions about which resources a student should access next. For instance, Chen et al. (2014) used collaborative filtering algorithms to generate candidate webpages and applied SPM to filter these candidates according to a learner's historical webpage visiting sequence. The remaining web pages were recommended to this learner. Other studies have applied SPM to resource accessing traces to understand how students learn the material. For example, Wong et al. (2019) applied SPM to sequences of video lectures watching events to check whether learners watched videos in the order that instructors planned.

### 6.2.1.4 Other data sources

SPM is applicable to data in other modes as long as the data can be formatted as a sequence of ordered itemsets. For example, Mudrick et al. (2019) used SPM to analyze students' eye-tracking data in a multimedia learning environment. In this study, an itemset was a fixation in a screen area, and a sequence was defined as a student's fixation trace. Knight and Martinez-Maldonado et al. (2017) explored how SPM could provide insights about rhetoric moves in students' written text. A sequence was defined as a student's written text, but an itemset could be defined at various levels, such as a sentence, a paragraph, and a section. McBroom et al. (2018) proposed a method that uses SPM to help teachers identify common mistakes in programming tasks. In this case, an itemset was defined as a submitted answer labeled with a code state (e.g., having syntax errors versus no syntax errors), and a sequence was composed of a student's submissions on a problem. Overall, few studies have explored how to apply SPM to educational data from these uncommon channels, but existing studies have shown the potential of SPM.

### *6.2.2 Research purposes*

With educational data in the above modes, SPM has been used in descriptive, relational, and predictive research for various purposes, including 1) mining learning behaviors, 2) analyzing and enriching educational theories, 3) evaluating the efficacy of interventions, 4) generating features for prediction and classification models, and 5) filtering learning resources for building recommender systems.

#### 6.2.2.1 Mining learning behaviors

Learning process data, such as event logs and discourse data, record detailed information about students' interactions with the learning environments, peers, and instructors. Frequent sequential patterns in such data indicate common behavior patterns across students (Zhou et al., 2010). These behavior patterns may reveal how students navigate their activities within a learning environment and inform how to update better the design of the learning experience (Mirzaei & Sahebi, 2019). For example, Kang et al. (2017) applied the cSPADE algorithm to gameplay logs in Alien Rescue, a serious game for teaching middle school students scientific problem-solving skills. The study aimed to analyze how sequential patterns of action may differ across multiple days of using the educational game. They observed how sequential patterns in the first few days represented exploration behaviors, while sequential patterns in the remaining days represented scientific problem-solving behaviors, including background search, generating and testing hypotheses, and constructing solutions. Emara et al. (2018) used SPM to compare the behaviors of a group of sixth-grade students working individually to another group working in pairs within Betty's Brain, a hypermedia learning environment teaching scientific phenomena. Differences in sequential patterns between the two groups suggested that students were better at fixing errors in solutions when they worked collaboratively than working individually.

#### 6.2.2.2 Enriching educational theories

Some studies have linked the interpretation of sequential patterns to educational theories. For instance, Taub and Azevedo (2018) applied SPM to investigate how hypothesis-testing behaviors and emotion levels together influenced learning and gameplay within Crystal Island, an educational game that teaches scientific inquiry skills and microbiology. In this game, students needed to collect and test various food items to find the cause of an illness. The hypothesis-testing behaviors were conceptualized as metacognitive monitoring strategies based on the information processing theory of SRL (Winne & Hadwin, 2008), while the emotion levels were conceptualized as the appraisals of events based on the component process model of emotions (Scherer, 2009). The results indicated that students who were less emotional and solved the task with one attempt showed more patterns of hypothesis-testing behaviors indicative of monitoring strategies, suggesting that they might be better at monitoring cognitive activities than the others. Thus, Taub and Azevedo's study enriched the understanding of the link between the information processing theory of SRL and the component process model of emotions. Kinnebrew and Segedy et al. (2017) argue that SPM may discover sequential patterns that do not match a theoretical problem-solving model, and these patterns represent new learning strategies used by students and can inform us about ways to refine the theoretical model.

### 6.2.2.3 Evaluating the efficacy of interventions

Studies have explored whether sequential learning behavior patterns could capture the effect of interventions. For example, Wong et al. (2019) designed weekly prompt videos to facilitate students to think about their plans, monitor, and reflect on learning in a Coursera course. They used SPM to compare students who watched at least one prompt video (prompt viewers) and those not watching any prompt video (non-viewers). The group of prompt viewers shared more sequential behavior patterns than non-viewers. In particular, prompt viewers tended to watch videos in the order that instructors planned.

### 6.2.2.4 Building predictive models

Sequential patterns from learning process data have been used to build predictive models. For example, Fatahi et al. (2018) used sequential patterns from students' event logs in Moodle to predict students' personality types, which were assessed via the Myers-Briggs Type Indicator (MBTI) questionnaire. The classification accuracies were 8% to 22% higher than chance. Jaber et al. (2016) argue that the position of a sequential pattern in students' event logs is important for classification tasks in education. Based on this idea, they proposed a SPM-based classification framework that segmented each sequence into $n$ bins with equal sizes. SPM was conducted within each bin, so a sequential pattern corresponded to n binary features, i.e., whether it appears in the first bin, second bin, and so on. The researchers applied the method to detect each student's role (executives, managers, or members) in collaborative projects. The SPM-based method had better performance than other methods in precision, recall, and F-1 score.

### 6.2.2.5 Developing educational recommender systems

Many studies have used SPM, together with other algorithms, to build learning resource recommender systems. A general procedure of using SPM for this purpose is the following: for a learner's resource accessing sequence $S = \{i_1, i_2, ..., i_n\}$, where $i_j$ is a learning resource, (1) in experts' (Fournier-Viger et al., 2010) or peers' (Chen et al., 2014) resource accessing sequences, discover frequent sequential patterns that contain subsequences identical or close to $S$; (2) resources that appear in these sequential patterns and are visited after $S$ are recommended to the learner (El-Ramly & Stroulia, 2004); (3) or these resources are combined with recommendations by other methods, such as collaborative filtering algorithms, to generate the final recommendation (Tarus et al., 2017).

Instead of directly building a recommender system, some studies have applied SPM to course enrollment data to advise students on course registration and schools on designing and refining professional programs' course paths. Such studies usually compare different academic performance groups' course enrollment paths and try to discover sequential course patterns frequent in the high-performing group but rare in the low-performing group. For example, Slim et al. (2016) applied SPM to electrical engineering undergraduates' course enrollment sequences. The result showed that students graduating with a high GPA tended to follow a course enrollment pattern distinct from low GPA students' course enrollment sequences.

## 6.3 How can SPM contribute to understanding the temporal aspects of learning?

The previous section discusses the potential of SPM in EDS. The current section discusses how to achieve this potential. Particularly, this section focuses on how SPM can assist us in understanding learning processes, including mining learning behaviors, enriching educational theories, and evaluating the efficacy of interventions. Sequential patterns are ordered itemsets and capture the temporal relationship between these itemsets. In education, this temporal relationship may reflect how learners arrange their activities or interact with peers, teachers, and learning environments. Knowledge and cognition are highly contextually dependent (Brown et al., 1989). Thus, a sequential pattern or temporal relationship in one setting may not generalize to another. Even if it does, the meaning is likely to be different. Nevertheless, studies have shown general ways in which SPM can contribute to the conceptual understanding of learning by discovering individuals' similarities, differences, and changes in learning behaviors.

This section uses studies on Betty's Brain, an open-ended learning environment, as examples to illustrate how SPM can be used to achieve these contributions. We focused on this environment because the Betty's Brain team regularly used SPM as a key method for their studies. As such, previous research on Betty's Brain has used SPM to achieve multiple research purposes related to understanding learning processes (i.e., mining learning behaviors, enriching educational theories, and evaluating the efficacy of interventions) and have used SPM in different ways to achieve these purposes (i.e., discovering similarities, differences, changes in learning behaviors). Additionally, two of the Betty's Brain studies were the first to combine SPM with statistical analyses for the purpose of discovering differences and changes in learning processes (Kinnebrew et al., 2013; Kinnebrew, Segedy et al., 2014). This section ends with a summary of the results of studies applying SPM in other learning environments.

### 6.3.1 An example: Combining SPM with statistical analyses to understand learning in an open-ended learning environment

Biswas and colleagues have conducted a series of studies that applied SPM to middle schoolers' event logs in Betty's Brain (Emara et al., 2018; Emara et al., 2017; Kinnebrew & Mack et al., 2014; Kinnebrew & Segedy et al., 2014, 2017; Kinnebrew et al., 2013). They have typically used the Pex-SPAM algorithm with a minimum support of 0.5 and a maximum gap of two to obtain frequent sequential patterns and applied statistical tests to these patterns to investigate differences and changes in students' learning processes. Before diving into these studies, it is helpful to introduce Betty's Brain.

#### 6.3.1.1 Betty's Brain

Betty's Brain is an open-ended computer-based learning environment (Biswas et al., 2016). Students learn about scientific phenomena, such as climate change, by teaching a virtual pedagogical agent, Betty. Students teach Betty by building a causal map of the scientific phenomenon, in which a causal (cause-and-effect) relationship is represented by a pair of concepts connected by a directed causal link (see the bottom right of Figure 6.1). To build this map, students can access hypermedia resource pages (Science Book in Figure 6.1) on relevant scientific concepts. Students can evaluate their causal modeling progress by asking Betty to take quizzes graded by a mentor agent, Mr. Davis. The top right of Figure 6.1 shows the questions, answers, and grades of a quiz. A gray grade means Betty could not answer the question because the question involved concepts or links that had not been added to the map. The student selected the second question, which was answered incorrectly, and the concepts and links that Betty used to answer this question (i.e., Betty's explanations) were highlighted in the causal map. Betty's quiz grades (correct and incorrect answers), along with her explanations, can help the student keep track of Betty's progress and, in turn, their own progress because

**Fig. 6.1** Screenshot of viewing quiz results and checking the chain of links Betty used to answer a quiz question.

Betty's correct and incorrect answers inform problems in the causal map. Students can then improve their understanding of the topic by re-reading the science book and tracking Betty's explanations to correct their perceived problems with their causal map.

Students' activities in Betty's Brain can be grouped into three large categories: (1) information seeking, including reading (read a page in the resources), search (search pages containing entered keywords), and note taking (create, view or edit a note); (2) solution construction, including edits (adding or deleting a concept, adding, deleting or modifying a causal link) and markings (marking a causal link as correct or incorrect); (3) solution assessment, including queries (ask Betty a cause-and-effect question based on the concepts and links on the map so far) and quizzes (assess the state of the map by having Betty take a quiz, viewing the quiz results and Betty's explanation on the results).

### 6.3.1.2 Discovering similarities in engagement and learning

Discovering similarities in learning processes is the most straightforward application of SPM in educational data because popular SPM algorithms, such as GSP, SPAM, and cSPADE, aim to find common sequential patterns across sequences. The discovered similarity may be among a group of learners or a single student's different learning sessions (Zhou et al., 2010). The similarity may reflect strategies common in students' learning (Moon & Liu, 2019), regardless of whether they are effective or ineffective.

In Betty's Brain, researchers have developed a task model which specifies strategies that students may use (Segedy et al., 2014). SPM has been used to discover candidate strategies that may increase the task model coverage. For instance, Kinnebrew and Segedy et al. (2017) found three meaningful sequential patterns that were not covered by the initial task model: *search -> search, note -> read -> note, and quiz -> adding link -> quiz*. *Search -> search* meant that students searched resource pages consecutively, suggesting that students might have difficulty finding a page on the first try. This pattern had a practical implication that the search module could be redesigned to help students quickly find relevant resource pages. *Note -> read -> note* meant that students switched between reading pages and taking notes frequently, indicating that students were tracking their understanding. *Quiz -> adding link -> quiz* meant that students added a link rather than reading resource pages after taking a quiz and checked the link correctness by taking another quiz. This pattern suggested that students might use a guess-and-check strategy.

### 6.3.1.3 Discover differences in engagement and learning

In addition to behaviors common among learners, researchers are also interested in identifying sequential patterns that differ across learners. This can be achieved via three steps:

1. Use a SPM algorithm to discover frequent candidate patterns within each group of learners.
2. Calculate frequency metrics of all candidate patterns for each group. This may include metrics such as the support value (the proportion of sequences that contain a sequential pattern; Agrawal & Srikant, 1995) or the instance value (the number of occurrences of a sequential pattern in a sequence; Lo et al., 2008).
3. Apply statistical tests to identify candidate patterns that occur with statistically significant different frequencies between groups. For support values of a sequential pattern, which are at the group level, researchers have used contingency table-based tests, such as Pearson chi-square tests, to examine differences in the support value (He et al., 2019). *t*-tests and the analysis of variances (ANOVA) have been used to examine instance differences (Kinnebrew et al., 2013), which are at the individual or sequence level. Studies in Betty's Brain mainly examined sequential patterns with different instance values between groups.

For example, Kinnebrew et al. (2013) used the Pex-SPAM algorithm in step 1 to obtain candidate patterns from students with high and low performance in Betty's Brain. Then, they used *t*-tests (step 3) to examine the two groups' differences in the instance values (step 2) of the candidate patterns. High-performing students had greater instance values in *adding an incoherent link -> quiz -> removing the incoherent link*. An incoherent link meant that students added this link without reading any relevant resource pages (possibly based on prior knowledge or guesses). This pattern indicates that high-performing students were more likely to take quizzes to check an incoherent link after adding it, and when the quiz result showed that the link was incorrect, they would remove it. In addition, the high-performing group was more likely to read pages that contained information about incorrect links in the map indicated by quiz results, while low-performing students tended to read irrelevant pages after viewing the quiz results. These results suggested that high-performing students more effectively used quizzes to track their understanding and navigate the subsequent activities.

Emara et al. (2017) let students work in pairs in Betty's Brain. Then, they grouped pairs into bothStudents or oneStudent groups based on the quality of their collaboration. Students in the bothStudents group showed many collaborative behaviors, such as elaborating on their partner's initiatives with an alternative, while students in the oneStudent group mainly worked individually. SPM revealed that the bothStudents group was more likely to read resource pages after quizzes than the oneStudent group. The bothStudents group tended to switch between editing, reading, and quizzes, while the oneStudent group tended to switch between editing and quizzes, suggesting that students in this group might solve the task with a trial-and-error strategy. Emara et al. (2018) further investigated how learning behavior differed when students worked individually and collaboratively in Betty's Brain. They let a group of students work individually while another students work in pairs. SPM found that, in comparison with the individual group, the collaboration group made the following patterns more often: (1) *adding a relevant and correct link -> quiz,* (2) *quiz -> removing incorrect links,* (3) *quiz -> read -> adding a relevant and correct link,* and (4) *read-> adding a relevant and correct link.* In contrast, the individual group tended to do multiple readings consecutively. These differences suggested that the collaboration group might be better at using quizzes to monitor their understanding and guide the next activity, and the individual group had difficulty in translating what they read into map construction.

Kinnebrew and Segedy et al. (2014) investigated the efficacy of scaffolding in Betty's Brain. Two groups of students used two versions of Betty's Brain with different scaffolding, while a control group used a version without scaffolding. In terms of the test scores, no treatment effect was found. However, SPM revealed sequential patterns that were used differently across groups. These differences were in line with the function of scaffolding that different groups received.

### 6.3.1.4 Discovering changes in engagement and learning over time

Investigating how sequential learning behavior patterns change over time can be achieved in five steps:
1. Use a SPM algorithm to discover frequent candidate patterns.
2. Break raw sequences into bins with equal sizes or break sequences based on a natural cutoff, such as learning sessions and days.

3. For each candidate pattern, calculate the metric of interest per learner per bin. As usual, the metric may be the support value or instance value.
4. For each candidate pattern, calculate a metric that characterizes its variation across bins. The metric may be information gain (Kinnebrew et al., 2013) or effect size (Zhang & Paquette, 2020).
5. Rank sequential patterns based on the metric computed in step 4. The top-ranked patterns have the largest variation over time.
6. Alternatively, researchers may omit steps 4 and 5 and apply repeated ANOVA to identify sequential patterns whose support or instance values statistically significantly differ across bins.

Kinnebrew and Segedy et al. (2014) segmented each student's event logs into five bins with an equal number of events and examined how the usage of frequent sequential patterns changed across bins. Results showed three clusters among these patterns in terms of the change of their usage: in cluster 1, the usage of sequential patterns decreased quickly over time; in cluster 2, the usage of sequential patterns decreased slowly; in cluster 3, the usage of sequential patterns increased quickly. Further analysis showed that the change of sequential patterns was in line with the scaffolding students received. For example, patterns representing knowledge construction behaviors were used steadily over time by students who received knowledge construction support, while the usage of such patterns decreased over time among students who received support for evaluating solutions or did not receive any support. Kinnebrew and Mack et al. (2014) found that the overall frequencies over time of some patterns were similar across groups that received different support, but these patterns showed significant differences across groups in the frequency change over time.

### 6.3.1.5 Summary

The above studies in Betty's Brain illustrate the application of SPM for various research purposes, such as extending the understanding of how students approach a learning task, evaluating the effect of scaffolding, and revealing how learners change behaviors over time. The ways that SPM achieves these purposes are summarized into three types: discovering individuals' similarities, differences, and changes in learning. The following section shows example studies where SPM achieves similar purposes in a wide range of learning environments beyond Betty's Brain.

### *6.3.2 Findings from applications of SPM in other learning environments*

Table 6.4 summarizes 20 example studies applying SPM to understand learning in environments other than Betty's Brain. A comprehensive review is out of the scope of this chapter, so we selected these studies from prior work. We searched prior work in ERIC and Engineering Village using terms related to (a) sequential pattern and (b) computer-supported learning. We used the AND operator and restricted that the terms must appear in the main text. The search was done in April 2022 and returned 209 articles. We excluded 24 articles that did not use sequential pattern mining in computer-supported learning. We selected the example studies from the remaining 189 articles. The selection is based on three criteria. First, the study used SPM to understand the temporal aspects of learning. Thus, we did not include studies that have used SPM only for predictive modeling or building recommender systems. Second, the study was conducted under a clear theoretical framework or topic. Third, we wanted to cover various topics and learning environments. Thus, we kept only one of the studies that shared the learning environment, topics, and the contributions of SPM (i.e., discovering individuals' similarities, differences, and changes in learning processes). Some of the selected studies have used analyses beyond SPM; however, we report only the results from the SPM analyses.

The context in these studies varied from computer-supported individual learning to collaborative learning and from simple learning environments, such as multimedia slides and online forums, to advanced technologies, such as educational games and ITS. The learning content mainly belonged to science, engineering, technology, and mathematics (STEM). Note this is not because SPM is only suitable for analyzing the STEM learning processes. The reason is perhaps that most educational technologies are designed for teaching STEM.

These studies have used SPM to explore various research topics, such as SRL, collaborative learning, scientific inquiry, at-risk students, knowledge building, and critical thinking. Among these topics, a dominant one is SRL. Some studies have used SRL as an external variable and investigated its relations with sequential patterns. For example, Sabourin et al. (2013) classified students as high, medium, and low SRL based on students' self-reported status during an educational game. They compared SRL groups' differences in behavioral patterns during the game. Some studies have coded raw events into SRL behaviors and analyzed sequential patterns of the SRL behaviors. For instance, Zheng et al. (2019) coded chat messages into eight categories of SRL behaviors. The others have mapped sequential patterns to SRL strategies. For example, Kia et al. (2020) linked sequential patterns to monitoring and planning strategies. We discuss why SPM is a popular method for the study of SRL in section 6.4.1.3.

Most of the studies have used SPM to understand differences in learning. This is unsurprising because two of the distinctive values of temporal analyses are identi-fying differences in learning processes and explaining variations in learning outcomes (Molenaar & Wise, 2022). The studies have grouped students by various factors, including students' backgrounds (e.g., prior domain knowledge and SRL proficiencies; Sabourin et al., 2013; Taub & Azevedo, 2019), experimental conditions (e.g., the information discrepancy type; Mudrick et al., 2019), during task behaviors (e.g., the use of SRL scaffolding and the collaboration levels; Martinez-Maldonado et al., 2013; Wong et al., 2019), and final performance (e.g., the problem-solving correctness and efficacy; Taub & Azevedo, 2018; Zhu et al., 2019). Note

Many studies have also used SPM to uncover similarities in learning behaviors. For example, Taub et al. (2018) found that learners showed common patterns of hypothesis-testing behaviors in learning scientific inquiry. Liu and Israel (2022) identified three problem-solving strategies from students' frequent sequential patterns during learning mathematics concepts and cognitive skills.

Two of the 20 studies have used SPM to investigate the changes in learning behaviors. Kang and Liu (2020) investigated how students' behavior patterns changed from day to day. They found no differences between the first and second days. But from the second to the fourth days, the number of frequent sequential patterns firstly decreased dramatically and then increased. Sun et al. (2022) compared high- and low-performing students' changes in behavior patterns over course phases. High-performing students' behavioral patterns became more complex from the initial stage to the stage of group learning and indicated group construction. By contrast, low-performing students' behavioral patterns in the initial stage were not distinct from the stage of group learning.

Note when investigating the differences and changes in sequential patterns, some studies have solely relied on descriptive statistics and qualitative observations. We do not recommend this practice because it makes evaluating the reliability of the results challenging. We suggest researchers use statistical tests to provide robustness information about the results whenever the statistical tests are applicable (see sections 6.3.1.3 and 6.3.1.4).

Overall, the 20 selected studies and research in Betty's Brain depict a broad picture of how to use SPM for EDS research and the potential of SPM in understanding learning processes. It is noteworthy that the use of SPM should be based on research context and purposes. Specifically, researchers need to consider the theoretical underpinnings of their studies when preprocessing the data, setting the parameters of a SPM algorithm, grouping learners for investigating differences in sequential patterns, choosing the temporal unit for investigating the change in patterns over time, and interpreting the sequential patterns.

**Table 6.4**. Summary of selected studies where SPM contributes to the understanding of learning

| | References | Learning context | Framework/ topic | Contributions of SPM | Main results based on SPM |
|---|---|---|---|---|---|
| 1. | Sabourin et al. (2013) | | SRL | Detecting differences | • High SRL students showed patterns indicative of recording information after receiving it more frequently than low SRL students. <br> • Low SRL students showed patterns indicative of random hypothesis testing behavior more often than high SRL students. <br> • Low SRL students' patterns suggested that they had difficulty in connecting different concepts. |
| 2. | Taub and Azevedo (2018) | Crystal Island, an educational game teaching microbiology | SRL; Component process model of emotion | Detecting differences | • Students at low-efficiency levels and high levels of facial expressions of emotions showed more sequential patterns representing less strategic hypothesis-testing behavior. <br> • Students at high-efficiency levels and both emotion levels had more sequential patterns indicative of strategic testing behavior. |
| 3. | Taub et al. (2018) | | SRL | Detecting similarities and differences | • Students showed common hypothesis-testing behavior patterns. <br> • Students solving the task more efficiently had fewer testing behavior patterns. <br> • Students with less efficiency had higher instance values in an ineffective behavior pattern. |

**Table 6.4**. Summary of selected studies …

| References | Learning context | Framework/ topic | Contributions of SPM | Main results based on SPM |
|---|---|---|---|---|
| 4. Taub and Azevedo (2019) | MetaTutor, ITS teaching the human circulatory system | SRL | Detecting differences | • High prior-knowledge students engaged in sequential patterns containing accurate metacognitive judgement, while low prior-knowledge students engaged in patterns with inaccurate metacognitive judgments.<br>• In both groups, few sequential patterns contained both cognitive and metacognitive events.<br>• High prior-knowledge students' sequential patterns contained more cognitive events than metacognitive events, while low prior-knowledge students' sequential patterns did not have any cognitive events. |
| 5. Martinez-Maldonado et al. (2013) | Cmate, a tabletop application for drawing concept maps | Collaborative learning; concept mapping | Detecting differences | • The more collaborative groups showed more sequential patterns that contained verbal discussions in conjunction with physical actions. In contrast, the less collaborative groups showed more sequential patterns that only had physical actions without speech.<br>• The less collaborative groups had more patterns where a student's brief speech did not get a response from the other members.<br>• The more collaborative groups showed more patterns that represented accessing solutions to trigger discussion. |

**Table 6.4**. Summary of selected studies …

| References | Learning context | Framework/ topic | Contributions of SPM | Main results based on SPM |
|---|---|---|---|---|
| 6. Jiang et al. (2015) | Virtual Performance Assessments for assessing science inquiry skills | Scientific inquiry; expert and novice | Detecting differences | • Novices in the learning environment showed more sequential patterns indicative of exploration behaviors than experienced students. |
| 7. Chen et al. (2017) | Knowledge Forum in primary school science courses | Knowledge building | Detecting differences | • In productive inquiry threads, students showed more sequential patterns indicative of (1) explaining after proposing a question, (2) questioning an explanation, (3) sustaining an explanation for questions, and (4) obtaining and analyzing information to propose, support, or improve an explanation.<br>• In improvable inquiry threads, students showed more sequential patterns of responding to explanations or new information with an opinion. |
| 8. Mudrick et al. (2019) | Twelve multimedia science content slides | Cognitive theory of multimedia learning | Detecting similarities and differences | • Learners showed some common sequential eye fixation patterns.<br>• Learners in different information discrepancy conditions showed differences in eye fixation patterns. |

**Table 6.4**. Summary of selected studies …

| References | Learning context | Framework/topic | Contributions of SPM | Main results based on SPM |
|---|---|---|---|---|
| 9. Kang et al. (2017) | Alien Rescue, a serious game for teaching middle schoolers scientific problem-solving skills | Scientific inquiry; expert and novice | Detecting similarities and differences | • The sequential patterns representing exploration behaviors were frequent in the first few days.<br>• Sequential patterns in the remaining days were indicative of scientific problem-solving behaviors.<br>• Low-performing students showed a larger number of different frequent sequential patterns than high-performing students. |
| 10. Kang and Liu (2020) | | Scientific inquiry; at-risk students | Detecting differences and changes | • On the first day, both at-risk and non-at-risk students showed many frequent sequential behavior patterns.<br>• From the second to the fourth days, the number of different frequent sequential patterns decreased dramatically in both groups and then increased.<br>• In the fifth and sixth days, the non-at-risk group had much more sequential patterns than the at-risk group. |
| 11. Zhu et al. (2019) | Teaching Teamwork, a computer-supported collaborative learning environment | Collaborative inquiry learning | Detecting similarities and differences | • In the successful condition, groups of learners had more regulation activity patterns than in the unsuccessful condition.<br>• In the successful condition, groups showed a sequential pattern indicative of maintaining a shared understanding of the task.<br>• In both conditions, groups engaged in a sequential pattern indicative of trial-and-error behaviors. |

**Table 6.4.** Summary of selected studies …

| References | Learning context | Framework/ topic | Contributions of SPM | Main results based on SPM |
|---|---|---|---|---|
| 12. Zheng et al. (2019) | Teaching Teamwork | SRL; SSRL | Detecting differences | • In the condition of successfully solving a task, students had more sequential patterns of regulation activities than in the unsuccessful condition.<br>• In the successful condition, groups tended to monitor progress after task analysis. |
| 13. Zheng et al. (2022) | BioWorld, a computer-based environment for practicing clinical reasoning skills | SRL; scientific inquiry | Detecting similarities and differences | • Efficient students focused on fewer clinical reasoning patterns than less efficient students<br>• Less efficient students showed disorganized clinical reasoning patterns |
| 14. Wong et al. (2019) | A course in Coursea teaching serious game | SRL | Detecting differences | • Learners who watched at least one SRL-prompt video showed more sequential patterns than those not watching any SRL-prompt video.<br>• The SRL-prompt viewers had a sequential pattern indicative of following the order of the videos presented in the MOOC. |
| 15. Kia et al. (2020) | MyLA, a learning dashboard embedded in Canvas | SRL | Detecting differences | • Low-performing students showed more sequential patterns than high-performing students.<br>• Sequential patterns indicative of monitoring and planning strategies were more frequent in high-SRL students than in low-SRL students. |

**Table 6.4**. Summary of selected studies …

| References | Learning context | Framework/ topic | Contributions of SPM | Main results based on SPM |
|---|---|---|---|---|
| 16. Malekian et al. (2020) | A MOOC course teaching discrete optimization | Assessment readiness | Detecting differences | • Students that failed an assessment showed sequential patterns of consecutive attempts on assessments and forum viewing.<br>• Students that passed an assessment showed sequential patterns of viewing and reviewing lectures. |
| 17. Chen and Wang (2020) | A web-based inquiry science environment | Inquiry-based learning | Detecting differences | • Low-performing students showed more sequential patterns of accessed course nodes in line with the designed learning sequence than high-performing students.<br>• High-performing students showed more sequential behavior patterns indicative of logical operation in a simulation experiment. |
| 18. Mishra et al. (2021) | ENaCT, a computer-based environment for learning critical thinking | Critical thinking | Detecting similarities | • Frequent patterns that contained different types of actions did not mark good critical thinking<br>• Investigating the semantic coherence between actions in a pattern may generate insights into critical thinking |

**Table 6.4**. Summary of selected studies …

| References | Learning context | Framework/ topic | Contributions of SPM | Main results based on SPM |
|---|---|---|---|---|
| 19. Liu and Israel (2022) | Zoombinis, a puzzle-based game for learning mathematics concepts and cognitive skills | Problem-solving in games | Detecting similarities and differences | • Students showed three common problem-solving strategies.<br>• Applying SPM within each problem-solving phase generated insights about how the usage of strategies differed across phases and which strategies might contribute to the transitions from lower to advanced phases. |
| 20. Sun et al. (2022) | A massive private online course teaching computer science | Online learning | Detecting differences and changes | • Action patterns differed across course phases<br>• High- and low-performing groups showed distinct action patterns |

## 6.4 The strengths, limitations, and future directions of SPM in EDS

Previous sections provided an overview of how to apply SPM for various research purposes. This section discusses the strengths and weaknesses of the application of SPM in EDS. In some cases, the strengths and weaknesses are built on the comparison with other sequential analysis methods. We highlight factors that lead to the weaknesses and limit the application of SPM and suggest research directions.

### 6.4.1 Strengths

Although educational researchers may be unfamiliar with SPM, it is easy to learn. As a sequential analysis method, it can reveal meaningful information about learning processes and is a powerful tool for investigating SRL. Compared with the other sequential analysis approaches, SPM is more flexible.

#### 6.4.1.1 Easy to learn and accessible

The rationale of SPM is straightforward: finding sequential patterns that frequently occur in a sequence dataset. Using SPM correctly does not require a detailed understanding of how SPM algorithms efficiently find the number of sequences containing a pattern and count pattern occurrences. Moreover, given the same parameter configuration, different algorithms discover the same sequential patterns (Fournier-Viger et al., 2017). Educational researchers simply need to know the meaning of the different parameters that can be used to adjust what types of patterns found by the algorithm. SPM tools are also accessible to educational researchers. SPM libraries are available in C++, Python, data mining tools such as WEKA and RapidMiner, and data analysis tools such as R. There is also a specialized tool, SPMF, which has a simple graphical user interface and is fast and lightweight (Fournier-Viger et al., 2016). It offers over 50 SPM algorithms

#### 6.4.1.2 Revealing meaningful information that may be ignored otherwise

SPM focuses on the temporal relationship between events. It can bring insights about learning that are different from those obtained using counts, proportions, and durations of individual events. Combining SPM with statistical inference methods may discover more differences between groups of learners than the analyses of individual events (Kinnebrew & Segedy et al., 2014; Martinez-Maldonado et al., 2013; Taub & Azevedo, 2018). For instance, a beforementioned study on Betty's Brain did not find a treatment effect on test scores but discovered meaningful differences in sequential patterns between groups receiving different treatments (Kinnebrew & Segedy et al., 2014). Similarly, Taub and Azevedo (2018) did not find statistically significant differences in the frequency of individual hypothesis-testing behaviors between groups at different performance and emotion levels, but they found interesting differences in sequential behavior patterns (see study 2 in Table 6.4).

### 6.4.1.3 A powerful tool for SRL research

SRL refers to the learners' process of actively managing and regulating their behavior, cognition, emotion, and motivation toward their learning goals (Zimmerman, 1990). SPM has been used as a powerful tool for investigating SRL because some operationalizations of SRL theories explicitly involve sequences of events unfolding over time.

For example, SRL events may be operationalized at three levels (Winne, 2010): the occurrence level, the contingency level, and the patterned contingency level. In this operationalization, the occurrence level considers the features of individual events, e.g., the frequency of taking a quiz. The contingency level considers the conditional probability of a subsequent event given a prior event, e.g., the probability of reading a resource page after taking a quiz. The patterned contingency level considers the arrangement of events that repeatedly occurs, e.g., the frequency of reading a book page after taking a quiz. Sequential patterns are at the patterned contingency level because they are temporally ordered events. Thus, SPM serves well the purpose of investigating SRL at the patterned contingency level. This is why nearly half (8) of the 20 studies in Table 6.4 are under the framework of SRL.

### 6.4.1.4 More flexible than other sequential analysis methods

Process mining and epistemic network analyses are also powerful EDS tools for understanding the temporal property of learning (Bannert et al., 2014; Paquette et al., 2021). However, these approaches aggregate individual sequences at group levels and depict the holistic learning process. In contrast, SPM can capture local learning patterns.

Another sequential analysis method, the lag-sequential analysis (LSA), also focuses on local patterns (Bakeman & Quera, 2011). However, LSA restricts that the gap between events of a sequential pattern must be fixed (in LSA, the gap is named lag). For example, when counting the occurrences of *quiz -> read*, if the gap is fixed to 1, reading a page immediately after taking a quiz is counted as an instance, but reading a page after making a note after taking a quiz is not. When the gap is fixed to 2, reading a page after making a note after taking a quiz is counted as an instance of *quiz -> read*, but reading a page immediately after taking a quiz is not. By contrast, SPM allows the gap to vary. Both the sequence of reading a page immediately after taking a quiz and the sequence of reading a page after making a note after taking a quiz can be counted as instances of *quiz -> read*. Similarly, the Markov chain model and transition metrics (Bosch & Paquette, 2021; Matayoshi & Karumbaiah, 2020) suffer the same issue as with LSA.

T-pattern analysis can also find frequent sequential patterns and does not suffer the issues of the above methods (Magnusson, 2000). However, T-pattern analysis requires that the dataset contains the timestamp of each event, and the interval between events must be meaningful. Many educational datasets, particularly event logs, do not meet this requirement. For example, a student may leave for a while after taking a quiz before reading a page. The interval between taking the quiz and reading is not the time that the student spends on the quiz.

### 6.4.2 Limitations and future directions

SPM is accessible and easy to learn, but it does not mean educational researchers can master this technique easily. Educational researchers may find various challenges when applying SPM to understand learning. Some challenges are shared by other sequence analysis methods, while others are unique to SPM.

#### 6.4.2.1 No available SPM libraries for computing instance values

Although various tools are available for conducting SPM, they usually compute the support values and do not return instance values. However, when applying SPM to learning process data, the support value may not be sufficient because it only counts whether a learner's event log contains a sequential pattern without considering how often a pattern is repeated. When a learner's event log contains hundreds or thousands of events, only using support value losses much information. For example, using only support value, a learner who does the *quiz -> read* pattern ten times while using a learning environment would be regarded as the same as a learner who does *quiz -> read* only one time.

Researchers have proposed efficient algorithms to compute instance values and share their programs (Wu et al., 2018; Wu et al., 2020). However, these programs were developed using C++ and Java code. Users need to know at least the basics of these programming languages in order to use these programs. Many educational researchers may not possess the necessary computer programming expertise. SPMF implements one algorithm that counts the instance value, but it only gives the sum of pattern instances across sequences. Users do not have access to pattern instances in each sequence, preventing further analysis, such as conducting statistical tests to examine whether a pattern occurs differentially between groups. This may be one factor that limits the application of SPM in education. Thus, a SPM application that is friendly to users without programming backgrounds and meets the unique needs of EDS researchers is worth developing.

### 6.4.2.2 Lack a guideline for preprocessing and parameter setting

Preprocessing is necessary for feeding event logs to SPM algorithms. As mentioned in section 6.1.1, educational researchers have used five ways to preprocess the data: filtering, collapsing, contextualizing, abstracting, and breaking. However, it is unclear how different preprocessing decisions would impact the results. For instance, should we collapse multiple *read* events? If so, should we collapse these events to *read* or *read-MULT* to distinguish a single read event from multiple read events? How would the results differ between decisions? These questions await future work.

Similarly, there is no clear guideline for parameter setting. What should the minimum support be to consider a pattern interesting? In other words, how many learners are required to show a sequential pattern for it to be safe to say a sequential pattern is frequent? What should the maximum gap between itemsets of a sequential pattern be? How would the results differ under different parameter configurations? And so on.

Unlike hyperparameters in machine learning models, which can be finely tuned via validation, researchers decide SPM parameters based on theories, prior studies, and experiences (Van Laer & Elen, 2018). Researchers can compare the results under different SPM parameters and choose values that produce the best result. However, without validation, the result may be overfitting the data. Moreover, the meaning of *best* is up to researchers. Researchers may choose parameters that produce the results they agree with most or that are in line with their hypotheses. Consequently, the results may be biased. A systematic guideline about preprocessing data, setting SPM parameters, and being transparent about these procedures will facilitate the application of SPM and benefit the field of EDS.

### 6.4.2.3 Excessive sequential patterns

SPM algorithms may generate excessive sequential patterns, most of which are uninteresting or irrelevant to the research purpose (Zhou et al., 2010). Educational researchers have applied interestingness metrics, such as lift and Jaccard coefficients, to rank sequential patterns and select the top-ranked for further analyses (Jiang et al., 2015; Kinnebrew & Killingsworth et al., 2017). Several studies have investigated the match between these metrics and expert judgment of interestingness on association rules in the field of education (Bazaldua et al., 2014; Merceron

& Yacef, 2008). To what extent these metrics can serve as evidence for the interestingness of sequential patterns is unclear. Besides, most interestingness metrics are based on support values and ignore the multiple occurrences of a sequential pattern within individual sequences. Only using support values may cause a lot of information loss. Interestingness metrics based on instance values may be more informative and produce sequential pattern rankings different from those based on support values.

Setting constraints, such as the maximum gap, the maximum pattern length, and the maximum window (which is the maximum gap between the first and last item of a pattern), also reduces the number of patterns. For example, a smaller maximum gap and maximum window lead to fewer patterns. However, as the interestingness metrics are based on support values, reducing the number of identified patterns by adjusting these parameters risks missing meaningful patterns and information loss. We recommend adjusting the parameters mainly based on theoretical considerations. For example, Liu and Israel (2022) set the minimum pattern length as three because a single activity typically came along with three events in the learning environment of their study.

Another option is to keep only generator patterns or closed patterns. A pattern is a generator if its support is not equal to any of its super-pattern, while a pattern is closed if its support is not equal to any sub-pattern (Fournier-Viger et al., 2017). Generator patterns and closed patterns are regarded as a concise and representative subset of all patterns, and thus, some algorithms only return either generator or closed patterns (Gao et al., 2008; Wang & Han, 2004). However, we argue that it is not a wise option to keep only one kind of pattern in education. A pattern may have the same support but different instance values with its super-patterns or sub-patterns. Moreover, although some super-patterns and sub-patterns have the same meaning (e.g., *watching a lecture video -> watching a lecture video -> taking a quiz* vs. *watching a lecture video -> taking a quiz*), others do not (e.g., *watching a lecture video -> watching a lecture video -> taking a quiz* vs. *watching a lecture video -> watching a lecture video*). In addition, the sub-pattern of an understandable pattern may be hard to interpret because of missing some critical events (Magnusson, 2000). In summary, there are a few solutions for addressing the issue of excessive patterns, but these solutions have various disadvantages in educational research. New approaches to reduce patterns in educational data are necessary.

### 6.4.2.4 Interpreting sequential pattern differences is challenging

Researchers must be cautious when interpreting differences in sequential patterns. First, when comparing two groups, if group A has more sequential patterns than group B, it does not necessarily mean that group A employs more strategies. For instance, in Zhu et al. (2019) study, the number of sequential patterns was greater when students successfully solved a task for those failing a task. The authors con-cluded that students used more strategies in the success condition than in the failure condition. However, students might be more engaged with the task and execute more events in the success condition than in the failure condition. The longer a student's event log is, the more sequential patterns the log might contain. Consequently, there were more sequential patterns in the success condition than in the failure condition.

Second, interpreting differences in the support and instance values of sequential patterns also entails the same caution. Student A having a higher instance value in *quiz -> read* than student B only means that student A uses this pattern more frequently than student B. It does not necessarily imply that student A is more likely to do reading after a quiz than student B. For example, assume that student A did a sequence of 200 actions in which *quiz -> read* appears four times, while student B did 100 actions in which *quiz -> read* appears twice. The instance value of *quiz -> read* for A is four, larger than that for B, i.e., two. However, it would be wrong to claim that A is more likely to use *quiz -> read* than B because if B also did 200 actions, the instance value might become four.

It seems that the issue might be solved by controlling for the sequence length. However, this would still not guarantee the claim that student A used this pattern more frequently than student B. The base rates of *quiz* and *read* also need to be controlled. Assume that students A and B both did a total of 100 actions in which *quiz -> read* appears four times for each student. However, student A did the *quiz* action ten times and *read* ten times, while Student B did *quiz* five times and *read* five times. In this case, the instance value of *quiz -> read* and the sequence length are the same for A and B. However, it would be wrong to claim that A and B used *quiz -> read* at the same frequency because the base rate of *quiz* and *read* are different for A and B.

Researchers may use other sequence analyses, such as LSA, to address these limitations. LSA characterizes a sequential pattern by the conditional probability that the second pattern event occurs given the first pattern event, with the simple probabilities of the second pattern event being controlled. However, as discussed in section 6.4.1.4, these metrics are only applicable when sequential patterns contain two events and the gap between events is fixed. Given the relative pros and cons of different sequence analyses, we follow the recommendation of prior studies: combining SPM and other sequence analyses may generate better insights about learning (Liu & Israel, 2022).

### 6.4.2.5 Sequential patterns do not imply causality

Sequential patterns are a set of temporally ordered events that co-occur regularly. Significant co-occurrences of events may suggest causality, but this kind of understanding of causality is not quite helpful (Reimann et al., 2014). When the context or the condition changes, the co-occurrence may disappear. Researchers should further investigate the mechanism behind the co-occurrence that explains why the previous events trigger the subsequent events (Reimann et al., 2014). This way, SPM can better contribute to understanding engagement and learning. Thus, educational researchers need to interpret the results of SPM based on proper learning theories if the purpose is to understand engagement and learning.

## 6.5 Conclusion

Sequential pattern mining (SPM) is a useful educational data science (EDS) method for uncovering the relative arrangement of learning events. It can be applied to common educational sequence datasets, such as event logs, discourse data, and resource-accessing traces. Through discovering similarities and differences in learners' activities and revealing the temporal change of learning behaviors, SPM can be used for mining learning behaviors, analyzing and enriching educational theories, evaluating the efficacy of instructional interventions, generating features for prediction models, and building educational recommender systems. Many SPM algorithms are publicly available and easy to use. As a sequential analysis method, SPM can reveal unique insights about learning processes and be powerful for SRL research. It is more flexible in capturing the relative arrangement of learning events than the other sequential analysis methods. Nevertheless, it is noteworthy that the use of SPM should be based on research context and purposes. Future research may improve the potential of SPM in EDS by developing tools for counting pattern occurrences as well as identifying and removing unreliable patterns. Another direction is to establish guidelines for data preprocessing, parameter setting, and interpretation of sequential patterns.

**Acknowledgments.** This research was partially funded by the China Scholarship Council (grant number 201806040180).

## 6.6 Appendix

### *6.6.1 R code for the synthetic example*

### install and load the R package that implements cSPADE

```
# install.packages('arulesSequences')
library(arulesSequences)
```

### Create the sequences. Each row of data contains an event, its event ID, and sequence ID.

```
data <- data.frame(
    sequenceID = c(rep(1, 6), rep(2,2), rep(3, 5), rep(4,
4)),
    events = c(
      "read", "hint", "attempt", "note", "attempt", "at-
tempt",
      "read", "attempt",
      "hint", "read", "attempt", "note", "attempt",
      "hint", "note", "read", "attempt")
    )
data$eventID <- 1:nrow(data)
```

### Convert the data to the basket format that cspade can handle
### Note that the first two columns should represent sequence ID and event ID

```
write.table(data[,c("sequenceID", "eventID", "events")],
            file = "formated_event_data.txt",
            sep=";",
            row.names = FALSE, col.names = FALSE,
            quote = FALSE)
data_baskets <- read_baskets("formated_event_data.txt",
                             sep = ";",
                             info=c("sequenceID","eventID"))
```

### Apply cspade to data_baskets

```
freq_patterns <- cspade(data_baskets,
                        parameter = list(support = 0.5,
                                         maxgap = 1))
```
### Inspect the frequent patterns. The minimum length of patterns in cspade is 1, so some patterns in freq_patterns are actually individual events.

```
inspect(freq_patterns)
```

### Convert the freq_patterns to a data.frame

```
freq_patterns_df <- as(freq_patterns, "data.frame")
```

### *6.6.2 Python code for the synthetic example*

### install and load the Python package that implements cSPADE

```
# !pip install Cython pycspade
from pycspade.helpers import spade, print_result
```

### Create a list to represent the sequences
### The first, second, and third columns are the sequence ID, event ID, and events, repsectively.
### At the time we wrote this example code, pycspade cannot handle events in string types.
### So we converted the events to integers: 1 - read, 2 - hint, 3 - attempt, 4 - note.

```
data = [
        [1,  1,     [1]],
        [1,  2,     [2]],
        [1,  3,     [3]],
        [1,  4,     [4]],
        [1,  5,     [3]],
        [1,  6,     [3]],
        [2,  7,     [1]],
        [2,  8,     [3]],
        [3,  9,     [2]],
        [3, 10,     [1]],
        [3, 11,     [3]],
        [3, 12,     [4]],
        [3, 13,     [3]],
        [4, 14,     [2]],
        [4, 15,     [4]],
        [4, 16,     [1]],
        [4, 17,     [3]]
    ]
```

### Apply cSPADE to the data

```
result = spade(data=data, support=0.5, maxgap = 1)
```

### Print the frequent patterns and interestingness measures

```
print_result(result)
```